\def\year{2021}\relax
\documentclass[letterpaper]{article} 
\usepackage{aaai21}  
\usepackage{times}  
\usepackage{helvet} 
\usepackage{courier}  
\usepackage[hyphens]{url}  
\usepackage{graphicx} 
\usepackage{natbib}  
\usepackage{caption} 
\usepackage{subcaption}
\usepackage{multirow}
\usepackage{amsmath,amssymb}
\usepackage{bm}
\def\x{$\times$}
\newlength\savewidth\newcommand\shline{\noalign{\global\savewidth\arrayrulewidth
  \global\arrayrulewidth 1pt}\hline\noalign{\global\arrayrulewidth\savewidth}}

\title{MVFNet: Multi-View Fusion Network for Efficient Video Recognition}
\author{
    Wenhao Wu\textsuperscript{\rm 1},
    Dongliang He\textsuperscript{\rm 1}\thanks{Corresponding author},
    Tianwei Lin\textsuperscript{\rm 1},
    Fu Li\textsuperscript{\rm 1},
    Chuang Gan\textsuperscript{\rm 2},
    Errui Ding\textsuperscript{\rm 1}\\
}
\affiliations{
    \textsuperscript{\rm 1} Department of Computer Vision Technology (VIS), Baidu Inc. \\
    \textsuperscript{\rm 2} MIT-IBM Watson AI Lab \\
    \{wuwenhao01, hedongliang01, lintianwei01, lifu, dingerrui\}@baidu.com, ganchuang1990@gmail.com

}

\begin{document}

\maketitle

\begin{abstract}
Conventionally, spatiotemporal modeling network and its complexity are the two most concentrated research topics in video action recognition. Existing state-of-the-art methods have achieved excellent accuracy regardless of the complexity meanwhile efficient spatiotemporal modeling solutions are slightly inferior in performance. In this paper, we attempt to acquire both efficiency and effectiveness simultaneously. First of all, besides traditionally treating $H\times W \times T$ video frames as space-time signal (viewing from the Height-Width spatial plane), we propose to also model video from the other two Height-Time and Width-Time planes, to capture the dynamics of video thoroughly. Secondly, our model is designed based on 2D CNN backbones and model complexity is well kept in mind by design.
Specifically, we introduce a novel multi-view fusion (MVF) module to exploit video dynamics using separable convolution for efficiency. It is a plug-and-play module and can be inserted into off-the-shelf 2D CNNs to form a simple yet effective model called MVFNet. Moreover, MVFNet can be thought of as a generalized video modeling framework and it can specialize to be existing methods such as C2D, SlowOnly, and TSM under different settings. Extensive experiments are conducted on popular benchmarks (\emph{i.e.}, Something-Something V1 \& V2, Kinetics, UCF-101, and HMDB-51) to show its superiority. 
The proposed MVFNet can achieve state-of-the-art performance but maintain 2D CNN's complexity.
\end{abstract}

\begin{figure}[t]
    \centering
    \includegraphics[width=0.48\textwidth]{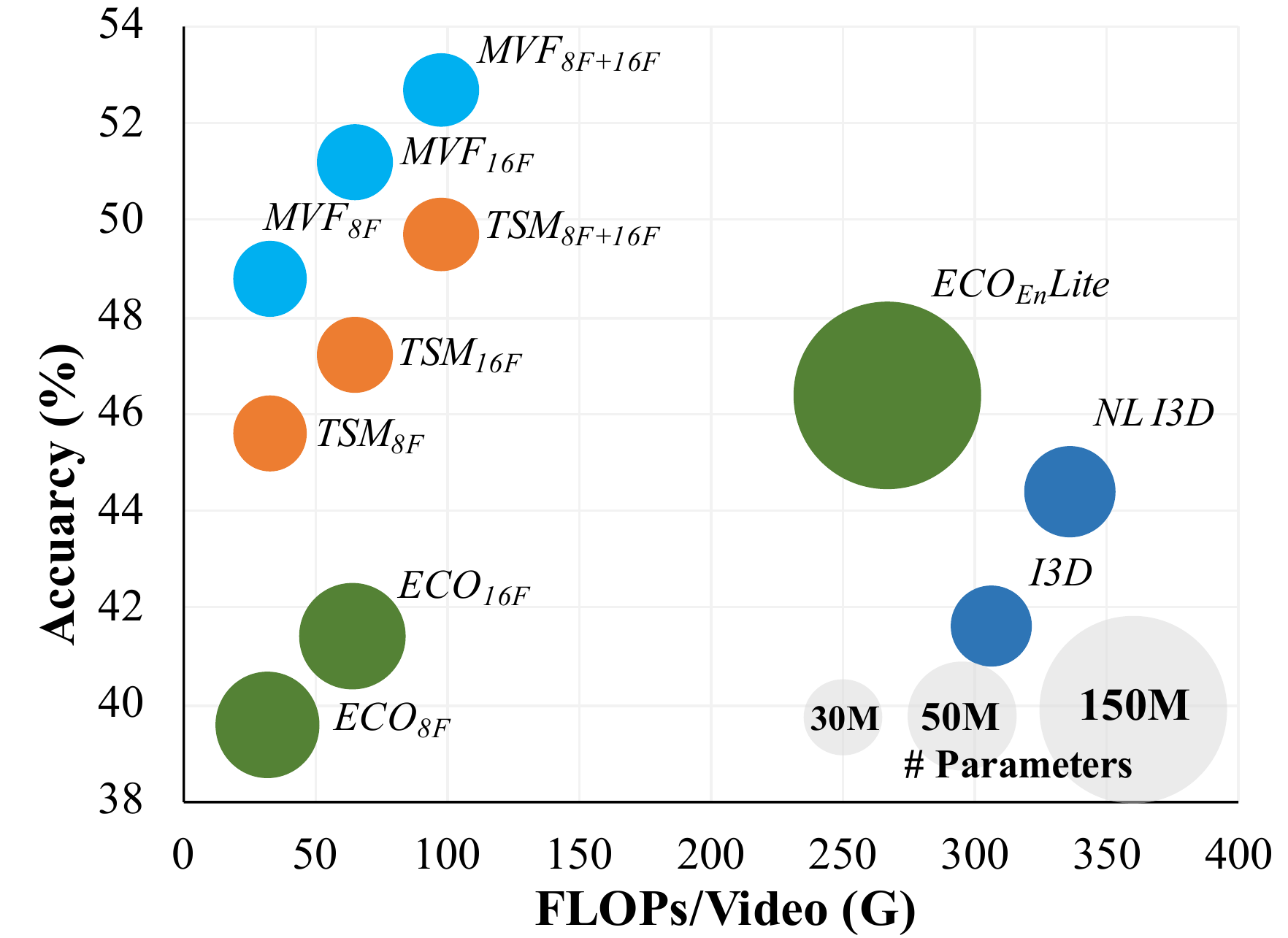}
    \caption{MVF achieves state-of-the-art performance on Something-Something V1 dataset and get better accuracy-computation trade-off than I3D, ECO and TSM. Figure credit: TSM~\cite{tsm}.}
    \label{fig:cmp_sth}
\end{figure}

\section{Introduction}
With the rapid development of the Internet and mobile devices, video data has exploded over the past years. Huge video information has far exceeded the processing capacity of the conventional manual system and attracted increased research interest in video understanding. Video recognition, as a fundamental task in video analytics, has become one of the most active research topics. It becomes increasingly demanding in a wide range of applications such as video surveillance, video retrieval, and personalized recommendation. Hence, recognition accuracy and inference complexity are of equal importance for the large scale applications. 

Recently, significant progress has been achieved in video action recognition following the deep convolutional network paradigm \cite{two-stream,tsn,i3d,nonlocal,tsm,slowfast}. 
Full-3D CNNs such as C3D~\cite{c3d} and I3D~\cite{i3d}, are intuitive spatial-temporal networks which are natural extension over their 2D counterparts to directly tackle 3D volumetric video data. Especially, the good performance achieved in I3D \cite{i3d} at the cost of thousands of GFLOPs\footnote{GFLOPs is short for float-point operations in Giga and is widely used to measure model complexity.}.  
In these methods, spatial and temporal features are jointly learned in an unconstrained way. However, the huge number of model parameters and computational burdens can largely limit the practicality of these methods. 

Then, some works~\cite{p3d,s3d,r2+1d} try to factorize 3D convolutional kernel into spatial (\emph{e.g.}, 1\x3\x3) and temporal part (\emph{e.g.}, 3\x1\x1) separately to reduce the cost. In practice, however, compared with their 2D counterparts, the increased computational overhead is still not negligible.
The recent state-of-the-art model TSM~\cite{tsm}, which achieves a good trade-off between performance and complexity, shifts feature along the temporal dimension instead of temporal convolution to model temporal dynamics in videos. TSM approximates spatiotemporal modeling done in 3D CNN while introduces zero FLOPs for the 2D CNN backbone. This inspires us to focus on designing efficient 2D CNN based architectures to learn more representative features for action recognition.

In this paper, we seek to design action recognition models to acquire both performance and efficiency. First of all, we propose to model dynamics in $H\times W \times T$ video signal from multiple viewpoints for performance improvement and we introduce an efficient spatiotemporal module, termed as \emph{Multi-View Fusion Module} (MVF Module). MVF is a novel plug-and-play module and can turn an existing 2D CNN into a powerful spatiotemporal feature extractor with minimal overhead. Specifically, the MVF module adopts three independent 1D channel-wise convolutions over the $T$, $H$ and $W$ dimensions respectively to capture multi-view information. To make this module more efficient, we decompose the feature maps into two parts, one acts as inputs of the three 1D channel-wise convolutions for multi-view spatiotemporal modeling and the other is directly concatenated with the outputs of the first part for making the original activation still accessible. 
In practice, we integrate the MVF module into the standard ResNet block to construct our MVF block. Then the final video architecture MVFNet is constructed by stacking multiple blocks. Interestingly, MVFNet can be regarded as the generalized spatiotemporal model and several existing methods such as C2D, SlowOnly~\cite{slowfast}, and TSM~\cite{tsm} can be specialized from MVFNet with different settings.

Extensive experimental results on multiple well-known datasets, including Kinetics-400 \cite{kay2017kinetics}, Something-Something V1 and V2 \cite{sth-sth}, UCF-101 \cite{ucf101} and HMDB-51 \cite{hmdb} show the superiority of our solution. As shown in Fig.~\ref{fig:cmp_sth}, MVFNet achieves excellent performance with quite limited overhead on Something-Something V1 and it is superior compared with existing state-of-the-art frameworks. The same conclusion can be drawn on other datasets. Codes and models are available\footnote{\url{https://github.com/whwu95/MVFNet}}.
Overall, our major contributions are summarized as follows:

\begin{itemize}
    \item Instead of only temporal modeling, we propose to exploit dynamic inside the three dimensional video signal from multiple viewpoints. A novel MVF module is designed to better exploit spatiotemporal dynamics. 
    \item The MVF module works in a plug-and-play way and can be integrated easily with existing 2D CNN backbones. Our MVFNet is a generalized video modeling network and it can specialize to become recent state-of-the-arts. 
    \item Extensive experiments on five public benchmark datasets demonstrate that the proposed MVFNet outperforms the state-of-the-art methods with computational cost (GFLOPs) comparable to 2D CNN.
\end{itemize}

\section{Related Work}

\textbf{2D CNNs} were extensively applied to conduct video recognition. Over the past years, inspired by the great success of deep convolution frameworks in image recognition~\cite{resnet,vgg,bn}, many methods have been proposed to explore the application of deep convolutional architectures on action recognition in videos. Among these methods, the Two-Stream architecture is a popular extension of 2D CNNs to handle video~\cite{two-stream,mv-cnn}, which can learn video representations respectively from RGB and optical flows or motion vector. To further boost performance, TSN~\cite{tsn} proposed a sparse temporal sampling strategy for the two-stream structure. TRN~\cite{trn} proposed by focusing on the multi-scale temporal relations among sampled frames. More recently, TSM~\cite{tsm}, STM~\cite{stm}, GST~\cite{GST}, GSM~\cite{GSM}, TEI~\cite{teinet}, TEA~\cite{li2020tea} perform efficient temporal modeling.

\begin{figure*}[t]
    \centering
    \includegraphics[width=0.90\textwidth]{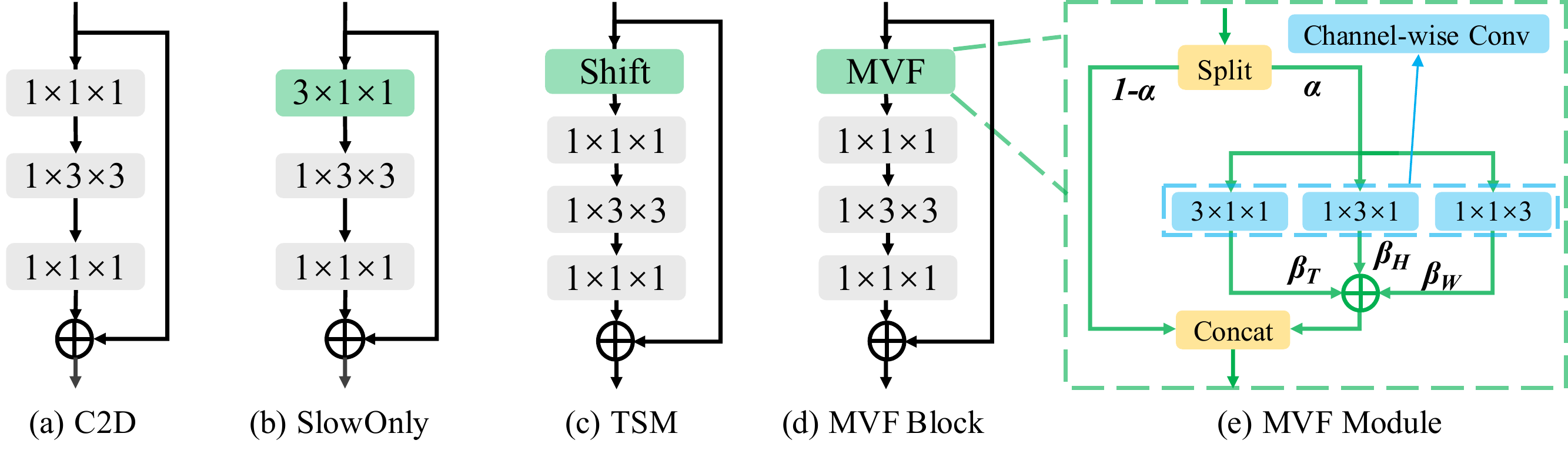}
    \caption{Illustration of various residual blocks for video action recognition. (a) shows a simple 2D ResNet-block. (b) shows the I3D$_{3\times1\times1}$ type block, which decouples the spatial and temporal filters by inflating the first 1\x1 kernel to 3\x1\x1. (c) shows a TSM block, which shifts partial feature maps along the temporal dimension for efficient temporal modeling. (d) shows our MVF block which integrate the MVF module into standard ResNet block. (e) depict the architecture of the MVF module.}
    \label{fig:mvf}
\end{figure*} 

\textbf{3D CNNs and (2+1)D CNN variants} based approach is another typical branch. C3D~\cite{c3d} is the first work in this line which directly learn spatiotemporal features from the video clip with 3D convolution. However, C3D has a huge number of parameters which makes it harder to train and more prone to over-fitting than 2D counterparts. To alleviate such problems, I3D~\cite{i3d} proposed to inflate the ImageNet pre-trained 2D convolution into 3D convolution for initialization. Following the I3D paradigm for spatiotemporal modeling, S3D \cite{s3d}, P3D \cite{p3d}, R(2+1)D \cite{r2+1d} and StNet \cite{stnet} are proposed to reduce computation overhead of 3D convolution while remaining the spatiotemporal modeling property. These (2+1)D CNN variants decompose 3D convolution into 2D spatial convolution followed by 1D temporal convolution on either per convolution operation basis or per 3D convolution network block basis. There exist several other networks that merge 2D and 3D information in CNN blocks to enhance the feature abstraction capability and resort to shallower backbones for efficiencies, such as ECO \cite{eco} and ARTNet \cite{ARN}. More recently, SlowFast~\cite{slowfast} explored the potential of different temporal speeds with two different 3D CNN architectures (\emph{e.g.}, Slow-Only and Fast-Only) to mimic two-stream fusion of 3D CNNs. In fact, our MVFNet can easily replace the Slow path in SlowFast.

The most closely related work to ours is CoST~\cite{Cost}, which also learns features from multiple views for action recognition. However, there are substantial differences between the CoST and the proposed MVF module. CoST learns collaborative spatiotemporal features through weight sharing among three regular 2D 3\x3 convolutions on $T$-$H$, $T$-$W$, and $H$-$W$ planes and replaces the middle 3\x3 convolution of ResNet block with CoST operation. CoST targets on performance and the three 2D 3\x3 convolutions are applied to the whole input feature map. On the contrary, we learn independent features via three different 1D channel-wise convolutions over $T$, $H$, and $W$ dimensions and collaborative learning does not work for our module. Besides of channel-wise 1D convolution, our MVF module performs multi-view modeling using part of the whole input feature map rather than the whole input feature map as done in CoST. Compared with CoST, our MVFNet achieves superior performance with much smaller computation cost.

There is also active research on dynamic inference~\cite{wu2020dynamic}, adaptive frame sampling techniques~\cite{wu2019adaframe,wu2019multi,korbar2019scsampler} and self-supervised video representation learning~\cite{chen2020RSPNet,han2020coclr}, which we think can be complementary to our approach.

\section{Approach}
\label{sec:approach}

In this section, first we present the technical details of our novel multi-view fusion (MVF) module which encodes spatiotemporal features effectively and efficiently with minimal overhead. Then we describe details of the MVFNet building block and how to insert this block into the off-the-shelf architecture of 2D CNNs. Finally, we present that MVF is a generalization of several existing methods such as C2D, SlowOnly~\cite{slowfast} and TSM~\cite{tsm}.

\subsection{Multi-View Fusion Module}
\label{subsec:mvf}
As shown in Fig~\ref{fig:mvf}, we compare the proposed method to several common competitive methods.  
C2D is a simple 2D baseline architecture without any temporal interaction before the last global average pooling.
A natural generalization of C2D is to turn it into a 3D convolutional counterpart. As 3D convolutions are computationally intensive, SlowOnly (the Slow path in SlowFast~\cite{slowfast}) inflated the first 1\x1 kernel in a residual block to 3\x1\x1 instead of inflating the 3\x3 kernel in a residual block to 3\x3\x3, and achieved promising performance.
To model temporal structure efficiently, TSM~\cite{tsm} involves some hand-crafted designs of shifting part of channels at each timestamp forward and backward along the temporal dimension, thus introduces zero FLOPs compared with C2D. 
An overview of our proposed multi-view fusion module is shown in Fig. 2(e), the input feature map channels are split into two parts, one for making part of the information in the original activation accessible, and the other for multi-view spatiotemporal modeling. In our module, multi-view modeling is performed via channel-wise convolution along the temporal, horizontal and vertical dimensions respectively, then the outputs of the three convolutions is element-wise added. Finally, the two part of features are concatenated together to combine the original activation and the multi-view modeling activation. 

Formally, let tensor $X\in \mathbb{R}^{C\times T\times H\times W}$ denote the input feature maps where $C$ is the number of input channels and $T$, $H$, $W$ are the temporal and spatial dimensions. Then let $X^1$, $X^2$ be the two splits of $X$ along the channel dimension where $X^1\in \mathbb{R}^{\alpha C\times T\times H\times W}$, $X^2\in \mathbb{R}^{(1-\alpha)C\times T\times H\times W}$ and $\alpha$ denoted the proportion of input channels for multi-view spatiotemporal features.
\begin{gather}
    O_T = \sum_i K^T_{c,i}\cdot X^1_{c,t+i,h,w}, \\
    O_H = \sum_i K^H_{c,i}\cdot X^1_{c,t,h+i,w}, \\
    O_W = \sum_i K^W_{c,i}\cdot X^1_{c,t,h,w+i},       
\end{gather}
where $c, t, x, y$ is the index of different dimensions of channel, time, height and width. $K^T, K^H, K^W$ is the channel-wise convolutional kernels used for modeling $H$-$W$, $W$-$T$and $H$-$T$ views respectively. In this paper, the kernel size of three channel-wise convolution is 3\x1\x1, 1\x3\x1 and 1\x1\x3. 
Then, the feature maps $O_T, O_H, O_W$ from three different views are fused
by a weighted summation as
\begin{equation}
    O^1 = \delta(\beta_T\cdot O_T +\beta_H\cdot O_H +\beta_W\cdot O_W),
\end{equation}
where $\delta$ is activation function, $O^1\in \mathbb{R}^{\alpha C\times T\times H\times W}$  is the activated feature maps, $\beta_T, \beta_H, \beta_W$ represents the weight value of the corresponding view. In this paper, we simply set $\beta_T=\beta_H=\beta_W=1$.
Finally, we get the MVF module output $Y\in \mathbb{R}^{C\times T\times H\times W}$ as
\begin{equation}
    Y = Concat(X^2, O^1),
\end{equation}
where $Concat$ represents concatenation operation along dimension of channel.

\subsection{Network Architecture}

We here describe how to instantiate the \emph{Multi-View Fusion Network} (MVFNet). The proposed MVF module is a flexible and efficient spatiotemporal module, which can be easily plugged into the existing 2D CNNs with a strong ability to learning the spatioltemporal features in videos. Therefore, our method is able to not only use the pre-trained ImageNet model for initialization to have faster training, but also bring very limited extra computation cost compared with 3D CNNs or (2+1)D CNNs. In practice, our MVFNet obviously improves the performance on different types of large-scale video datasets: scene-related one such as Kinetics-400, and temporal-related one such as Something-Something. 

We observed that many recent state-of-the-art action recognition methods~\cite{tsm,slowfast,Cost,teinet} usually use ResNet as the backbone network due to its simple and modular structure. For a fair comparison with these state-of-the-arts, in our experiments, we instantiate the MVFNet using ResNet as the backbone. Specifically, we integrate the proposed module into standard ResNet block to construct our MVF block then form our MVFNet. The overall design of the MVF block is illustrated in Fig.~2(d), MVF module is inserted before the first convolution of ResNet Block.

Unless specified, in our experiments we choose the 2D ResNet-50 as our backbone for its trade-off between the accuracy and speed.
Following recent common practice~\cite{tsm}, there are no temporal downsampling operations in this instantiation. The ResNet backbone is initialized from ImageNet pre-trained weights and our MVF modules do not need any specialized initialization strategy, simple random gaussian initialization work pretty well.

\subsection{Relation to Existing Methods}
Here we discuss the connection between MVF and other methods shown in Fig.~\ref{fig:mvf}. Details are described in the following:
\begin{itemize}
    \item With $\alpha$=0, MVF module degenerates to C2D which only focus on learning the spatial feature representation.
    \item With $\alpha$=1 and $\beta_H$=$\beta_W$=0, MVF module collapses to SlowOnly~\cite{slowfast} with depth-wise separable convolution. Put simply, the first 1\x1\x1 convolution of ResNet block perform as a point-wise convolution, then the 3\x1\x1 channel-wise convolution followed by the 1\x1\x1 point-wise convolution naturally form the depth-wise separable convolution.
    \item With $\alpha$=1/4 and $\beta_H$=$\beta_W$=0, MVF module could be viewed as a learnable TSM~\cite{tsm}. More specifically, in fact TSM could be viewed as a channel-wise 3\x1\x1 convolution, where convolution kernel is fixed as [0, 0, 1] for forward shift, and [1, 0, 0] for backward shift.
\end{itemize}

Above all, we conclude that MVFNet can be regarded as the generalized spatiotemporal model and above approaches can be specialized from MVFNet with different settings.

\section{Experiments}
\subsection{Datasets and Evaluation Metrics}
We evaluate our method on three large-scale video recognition benchmarks, including Kinetics-400 (K400)~\cite{kay2017kinetics}, Something-Something (Sth-Sth) V1\&V2 ~\cite{sth-sth}, and other two small-scale datasets, UCF-101~\cite{ucf101} and HMDB-51~\cite{hmdb}. Kinetics-400 contains 400 human action categories and provides around 240k training videos and 20k validation videos. 
For Something-Something datasets, the actions therein mainly include object-object and human-object interactions, which require strong temporal relation to well categorizing them. V1 includes about 110k videos and V2 includes 220k video clips for 174 fine-grained classes.
To elaborately study the effectiveness of our method on these two types of datasets, we use Kinetics-400 and Sth-Sth V1 for the ablation experiments. Moreover, transfer learning experiments on the UCF-101 and HMDB-51, which are much smaller than Kinetics and sth-sth, is carried out to show the transfer capability of our solution.

We report top-1 and top-5 accuracy (\%) for Kinetics and top-1 accuracy (\%) for Something-Something V1\&V2.
For UCF-101 and HMDB-51, we follow the original evaluation scheme using mean class accuracy. Also, we report the computational cost (in FLOPs) as well as the number of model parameters to depict model complexity. In this paper, we only use the RGB frames of these datasets for experiments.

\subsection{Implementation Details}
\label{sec:imp}

\begin{table*}
		\begin{subtable}[t]{0.45\textwidth}
		\centering\footnotesize
		\setlength{\tabcolsep}{2.0pt}
		\renewcommand{\arraystretch}{1.05}
			\begin{tabular}{l|cccc|cccc}
			\shline
			\multirow{2}{*}{Setting}  & \multicolumn{4}{c|}{Sth-sth v1} &  \multicolumn{4}{c}{Kinetics-400} \\ \cline{2-5}  \cline{6-9} 
		   & \#F & Top-1       & Top-5 &FLOPs   & \#F   & Top-1       & Top-5  & FLOPs    \\ \hline
				$\alpha$=0 & 8 & 17.12 & 43.46 & 32.88G & 4 & 71.87 & 90.02 & 16.44G \\
				$\alpha$=1/8 & 8 & 49.74 & 78.09 & 32.90G & 4 & \textbf{74.21} & 91.34 & 16.45G \\
				$\alpha$=1/4 & 8 & 49.24 & 77.91 & 32.92G & 4 & 74.18 & \textbf{91.46} & 16.46G \\
				$\alpha$=1/2 & 8 & \textbf{50.48} & \textbf{79.14} & 32.96G & 4 & 74.21 & 91.42 & 16.48G \\
				$\alpha$=1 & 8 & 49.73 & 77.94 & 33.04G & 4 & 73.75 & 91.40 & 16.52G \\ \shline
			\end{tabular}
			\caption{\textbf{Parameter choices of $\bm{\alpha}$}. Backbone: R-50.}
			\label{tab:ablation:ratio}
		\end{subtable}
		\begin{subtable}[t]{0.55\textwidth}
		\centering\footnotesize
		\setlength{\tabcolsep}{2.0pt}
		\renewcommand{\arraystretch}{1.05}
			\begin{tabular}{l|c|cccc|cccc} 
			\shline
			\multirow{2}{*}{Stages}  & \multirow{2}{*}{Blocks} & \multicolumn{4}{c|}{Sth-sth v1, \textbf{$\bm{\alpha}$=1/2}} &  \multicolumn{4}{c}{Kinetics-400, \textbf{$\bm{\alpha}$=1/8}} \\ \cline{3-6}  \cline{7-10} 
			&  & \#F  & Top-1     & Top-5 & FLOPs  & \#F    & Top-1       & Top-5 & FLOPs     \\ \hline
				None & 0 & 8 & 17.12 & 43.46 & 32.88G & 4 & 71.87 & 90.02 & 16.44G \\
				res\{5\} & 3 & 8 & 46.02 & 75.60 & 32.90G & 4 & 73.46 & 91.09 & 16.44G \\
				res\{4,5\} & 9 & 8 & \textbf{50.48} & \textbf{79.14} & 32.96G & 4 & 74.21 & 91.34 & 16.45G \\
				res\{3,4,5\} & 13 & 8 & 49.72 & 78.82 & 33.04G  & 4 & 74.08 & 91.51 & 16.46G \\
				res\{2,3,4,5\} & 16 & 8 & 49.95 & 77.96 & 33.12G  & 4 & \textbf{74.22} & \textbf{91.56} & 16.47G \\ \shline
			\end{tabular}    
			\caption{\textbf{The number of MVF Blocks} inserted into R-50.}
			\label{tab:ablation:stage}
		\end{subtable}
		\\[7pt]
		\begin{subtable}[t]{0.24\textwidth}
		\centering\footnotesize
		\setlength{\tabcolsep}{1.0pt}
		\renewcommand{\arraystretch}{1.05}
			\begin{tabular}{l|cc|cc}
			\shline
				\multirow{2}{*}{Views}   & \multicolumn{2}{c|}{Sth v1} &  \multicolumn{2}{c}{K400}  \\ \cline{2-3}  \cline{4-5}
			& \#F & Top-1      & \#F  & Top-1    \\ \hline
				T  & 8 & 49.13  & 4 & 73.72  \\
				T-H & 8 & 49.22  & 4 &  74.01  \\
				T-W  & 8 & 49.31 & 4 & 73.88 \\
				T-H-W & 8  & \textbf{50.48} & 4 & \textbf{74.21}  \\
				T-H-W (S) & 8 & 47.21 & 4 & 73.81  \\ \shline
			\end{tabular}	   
			\caption{\textbf{Study on the different views of MVF module}. Backbone: R-50. S denotes weight  sharing.}
			\label{tab:ablation:view}
		\end{subtable}   
		\hspace{2mm}
		\begin{subtable}[t]{0.28\textwidth}
		\centering\footnotesize
		\setlength{\tabcolsep}{1.0pt}
		\renewcommand{\arraystretch}{1.05}
			\begin{tabular}{l|c|c|cc}
			\shline
			\multirow{2}{*}{Method}   & \multicolumn{1}{c|}{Sth v1} &  \multicolumn{1}{c|}{K400} & \multirow{2}{*}{FLOPs} & \multirow{2}{*}{Params} \\ \cline{2-3} 
				& Top-1         & Top-1         &  & \\ \hline
				C2D  & 17.1 & 71.4  & 32.9G & 24.3M \\
				TSM   & 47.2 & 74.1 & 32.9G & 24.3M \\
				SlowOnly  & -  & 74.9 & 41.9G & 32.4M\\ 
				CoST$^*$ & -  & - & 45.8G & 24.3M \\ \hline
				MVFNet & \textbf{50.5}  & \textbf{76.0} & 32.9G & 24.3M \\
				\shline
				
			\end{tabular}	   
			\caption{\textbf{Study on the effectiveness of MVFNet}. Backbone: R-50, 8f input. * indicates our implementation.}
			\label{tab:ablation:compare}	
		\end{subtable}
		\hspace{2mm}
		\begin{subtable}[t]{0.19\textwidth}
		\centering\footnotesize
		\setlength{\tabcolsep}{1.0pt}
		\renewcommand{\arraystretch}{1.05}
			\begin{tabular}{l|ccc}
			\shline
				\multicolumn{1}{l|}{} & \#F & Top-1  & FLOPs  \\
				\hline
				\multirow{3}{*}{R-50} & 4 &  74.21  & 16.45G   \\
				& 8 &  75.99 &  32.90G  \\
				& 16 &  77.04 &  65.81G  \\
				\hline
				\multirow{3}{*}{R-101}  & 4 & 75.98 &  31.36G  \\
				& 8  & 77.46 &  62.72G  \\
				& 16 & 78.42 &  125.45G \\ \shline
			\end{tabular}   
			\caption{\textbf{Advanced backbones for MVFNet on Kinetics-400}.}
			\label{tab:ablation:various}
		\end{subtable}
		\hspace{2mm}
		\begin{subtable}[t]{0.226\textwidth}
		\centering\footnotesize
		\setlength{\tabcolsep}{1.0pt}
		\renewcommand{\arraystretch}{1.05}
			\begin{tabular}{l|c|cc}
			\shline
				\multicolumn{1}{l|}{} & Model & Top-1  & FLOPs  \\
				\hline
				\multirow{2}{*}{Mb-V2} & C2D &  64.4  & 1.25G   \\
				& MVF &  67.5 &  1.25G  \\
				\hline
				\multirow{2}{*}{R-50}  & C2D & 71.9 &  16.44G  \\
				& MVF  & 74.2 &  16.48G  \\  \shline
				
				\multicolumn{4}{c}{} \\ 
				\multicolumn{4}{c}{} \\ 
			\end{tabular}    
			\caption{\textbf{Different backbones for MVFNet on Kinetics-400}. Mb-V2 denotes MobileNet-V2.}
			\label{tab:ablation:mbv2}
		\end{subtable}  		
		
		\caption{{Ablation studies} on \textbf{Something-Something V1} and \textbf{Kinetics-400}. We show top-1 and top-5 classification accuracy (\%), as well as computational complexity measured in FLOPs (floating-point operations) for a  single clip input of spatial size $224^2$. \#F indicates the number of frames sampled from each video clip. We follow the common setting to sample multiple clips per video (10 for Kinetics-400, 2 for Something-Something V1).
		}
		\label{tab:ablations}
	\end{table*}

\textbf{Training.} 
We utilize 2D ResNet~\cite{resnet} as our backbone and train the model in an end-to-end manner. We use random initialization for our MVF module. Following the similar practice in \cite{nonlocal} on Kinetics-400, we sample frames from a set of consecutive 64 frames per video. For Something-Something V1 \& V2, we observe that the duration of most videos normally has less than 64 frames, thus we employ the similar uniform sampling strategy to TSN~\cite{tsn} to train our model. In our experiments, we sample 4, 8 or 16 frames as a clip. The size of the short side of these frames is fixed to 256 and then random scaling is utilized for data augmentation. Finally, we resize the cropped regions to 224\x224 for network training. 

On the Kinetics-400 dataset, the learning rate is 0.01 and will be reduced by a factor of 10 at 90 and 130 epochs (150 epochs in total) respectively. For Something-Something V1 \& V2 dataset, our model is trained for 50 epochs starting with a learning rate 0.01 and reducing it by a factor of 10 at 30, 40 and 45 epochs. For these large-scale datasets, our models are initialized by pre-trained models on ImageNet~\cite{deng2009imagenet}. For UCF-101 and HMDB-51, we followed the common practice to fine-tune from Kinetics pre-trained weights and start training with a learning rate of 0.01 for 25 epochs. The learning rate is decayed by a factor 10 every 10 epochs. For all of our experiments, we utilize SGD with momentum 0.9 and weight decay of 1e-4 to train our models on 8 GPUs. Each GPU processes a mini-batch of 8 video clips by default. When changing to a larger batch size b of each GPU for higher gpu memory usage, we use linear scaling initial learning rate (0.01\x b/8).

\textbf{Inference.} 
Following the widely used practice in~\cite{tsm,teinet}, two ways for inference are considered to trade-off accuracy and speed. (a) For high accuracy, we follow the common setting in \cite{nonlocal,slowfast} to uniformly sample multiple clips from a video along its temporal axis. We sample 10 clips for Kinetics-400 and 2 clips for others. For each clip, we resize the short side to 256 pixels and take 3 spatial crops in each frame. Finally, we average the softmax probabilities of all clips as the final prediction.
(b) For efficiency, we only use 1 clip per video and a central region of size 224\x224 is cropped for evaluation.

\subsection{Ablation Studies}
To comprehensively evaluate our proposed MVF module, in this section we provide ablation studies on both Kinetics-400 and Something-Something V1 datasets which represent the two types of datasets. 
Table~\ref{tab:ablations} shows a series of ablations. Accordingly, the effectiveness of each component in our framework is analyzed as follows.

\textbf{Paramter Choice.} 
As shown in Table~\ref{tab:ablation:ratio}, we compare networks with different  proportion of input channels ($\alpha=0, 1/8, 1/4, 1/2, 1$) for multi-view spatiotemporal feature. Here we add MVF blocks into res$_{4-5}$ for efficiency. Especially, when $\alpha=0$, MVFNet becomes exactly C2D. Our approach achieves considerable absolute improvement over C2D baseline on both datasets (+33.36\% for Sth-Sth v1, +2.34\% for Kinetics-400), which demonstrates the effectiveness of the MVF blocks. For Kinetics-400, we observe that the change in $\alpha=1/8, 1/4, 1/2$ appeared to have little impact on performance thus we choose $\alpha=1/8$ for efficiency in the following experiments. As for Something-Something V1, our method with $\alpha=1/2$ achieves the highest Top-1 accuracy compared with the other settings, so 1/2 is adopted in the following experiments.

\begin{table*}[th]
\centering
\scalebox{0.93}{
\begin{tabular}{ccccccc}
\shline
\bfseries Method & \bfseries Backbone & \bfseries Frames\x \bfseries Crops\x \bfseries Clips & \bfseries GFLOPs & \bfseries Top-1 & \bfseries Top-5\\
\hline
I3D~(Carreira et al. 2017) & Inception V1 & 64\x N/A\x N/A & 108\x N/A & 72.1\% & 90.3\%\\
S3D-G~\cite{s3d} & Inception V1 & 64\x3\x10 & 71.4\x30 & 74.7\% & 93.4\% \\
TSN~\cite{tsn} & Inception V3 & 25\x10\x1  & 80\x10 & 72.5\% & 90.2\% \\
ECO-RGB$_{En}$~(Zolfaghari et al. 2018) & BNIncep+Res3D-18 & 92\x1\x1 & 267\x1 & 70.0\% & -\%\\
R(2+1)D~\cite{r2+1d} & ResNet-34 & 32\x1\x10 & 152\x10 & 74.3\% & 91.4\%\\ 
X3D-M~\cite{feichtenhofer2020x3d} & - & 16\x3\x10 & 6.2\x30 & 76.0\% & 92.3\% \\
 \hline
 
STM~\cite{stm} & ResNet-50& 16\x3\x10 & 67\x30 & 73.7\% & 91.6\% \\ 

TSM~\cite{tsm} & ResNet-50 & 8\x3\x10 & 33\x30 & 74.1\% & 91.2\%\\
SlowOnly~\cite{slowfast} & ResNet-50 & 8\x3\x10 & 41.9\x30 & 74.9\% & 91.5\% \\ 
TEINet~\cite{teinet} & ResNet-50& 8\x3\x10 & 33\x30 & 74.9\% & 91.8\% \\ 
TEA~\cite{li2020tea} & ResNet-50& 8\x3\x10 & 33\x30 & 75.0\% & 91.8\% \\  
Slowfast~\cite{slowfast} & R50+R50 & (4+32)\x3\x10 & 36.1\x30 & 75.6\% & 92.1\% \\
NL+I3D~\cite{nonlocal} & ResNet-50 & 32\x3\x10 & 70.5\x30 & 74.9\% & 91.6\% \\
NL+I3D~\cite{nonlocal} & ResNet-50 & 128\x3\x10 & 282\x30 & 76.5\% & 92.6\% \\ \hline

MVFNet & ResNet-50& 8\x3\x10 & 32.9\x30 & \textbf{76.0}\% & \textbf{92.4}\% \\
MVFNet & ResNet-50& 16\x3\x10 & 65.8\x30 & \textbf{77.0}\% & \textbf{92.8}\% \\ \hline\hline

ip-CSN~\cite{CSN} & ResNet-101 & 32\x3\x10 & 82\x30 & 76.7\% & 92.3\% \\
SmallBig~\cite{li2020smallbignet} & ResNet-101 & 32\x3\x4 & 418\x12 & 77.4\% & 93.3\% \\
SlowOnly~\cite{slowfast} & ResNet-101 & 16\x3\x10 & 185$\times$30 & 77.2\% & -\% \\
NL+I3D~\cite{nonlocal} & ResNet-101 & 128\x3\x10 & 359\x30 & 77.7\% & 93.3\% \\ 
Slowfast~\cite{slowfast} & R101+R101 & (8+32)\x3\x10 & 106$\times$30 & 77.9\% & 93.2\% \\
Slowfast~\cite{slowfast} & R101+R101 & (16+64)\x3\x10 & 213$\times$30 & \textbf{78.9\%} & \textbf{93.5\%} \\
TPN~\cite{tpn} & ResNet-101 & 32\x3\x10 & 374$\times$30 & \textbf{78.9\%} & \textbf{93.9\%} \\
\hline
MVFNet & ResNet-101 & 8\x3\x10 & 62.7\x30 & 77.4\% & 92.9\% \\
MVFNet & ResNet-101 & 16\x3\x10 & 125.4\x30 & 78.4\% & 93.4\% \\ 
MVFNet$_{En}$ & R101+R101 & (16+8)\x3\x10 & 188.1\x30 & \textbf{79.1}\% & \textbf{93.8}\%  \\ 

\shline

\end{tabular}
}
\caption{Comparison with the state-of-the-art models on Kinetics-400. Similar to \cite{slowfast}, we report the inference cost by computing the GFLOPs (of a single view) $\times$ the number of views (temporal clips with spatial crops). N/A denotes the numbers are not available for us.}
\label{tab:sota_kinetics}
\end{table*}

\textbf{The Number of MVF Blocks.} 
We denote the conv2\_x to conv5\_x of ResNet architecture as res$_2$ to res$_5$. To figure out how many MVF blocks can obtain a good trade-off, we gradually add MVF blocks from res$_5$ to res$_2$ in ResNet-50. 
According to the results in Table~\ref{tab:ablation:ratio}, we set $\alpha$ to $1/2$ for Something-something V1 and $1/8$ for Kinetics-400.
As shown in Table~\ref{tab:ablation:stage}, on Kinetics-400, the improvement brought by MVF blocks on res$_{4-5}$, res$_{3-5}$ or res$_{2-5}$ is comparable. res$_{2-5}$ outperforms res$_{4-5}$ by 0.01\% but 7 extra MVF-blocks are needed. On Something-something V1, MVF-blocks added to res$_{4-5}$ achieves the highest top-1 accuracy. Thus we use MVF block in stages of res$_{4-5}$ in the following experiments for efficiency.

\textbf{Different Views.} 
We also evaluate impacts of different views in MVF module. As illustrated in Fig. 2(e), we can get different multi-view fusion by controlling $\beta_T, \beta_H$ and $\beta_W$. For example, we simply set $\beta_T$=$\beta_H$=$\beta_W$=1 to get view of $T$-$H$-$W$. Also, typical $T$-$H$, $T$-$W$, $T$ views can be easily obtained. From Table~\ref{tab:ablation:view}, we can see that $T$-$H$-$W$ outperforms $T$ by 1.35\% and 0.49\% on Something-Something V1 and Kinetics-400, respectively. Moreover, we also try the collaborative feature learning in CoST~\cite{Cost} by sharing the he convolution kernels among different views. However, with weight sharing among different views, accuracy gets degraded by 3.2\% and 0.4\% on two datasets, respectively.

\textbf{Comparison with Other Temproal Modules.} 
Here we make a comparison with methods described in Sec.~\ref{subsec:mvf} under the same setting of backbone and inputs. 
We list FLOPs and the number of parameter for all models in Table~\ref{tab:ablation:compare}, our MVFNet is more lightweight (32.9G \emph{vs.} 41.9G) than the SlowOnly branch of SlowFast~\cite{slowfast} and achieves a better accuracy than it (\textbf{76.0}\% \emph{vs.} 74.9\%). 
Also, our MVFNet outperforms the TSM~\cite{tsm} with a large margin on both datasets (Sth V1: \textbf{50.5}\% \emph{vs.} 47.2\%, K400: \textbf{76.0}\% \emph{vs.} 74.1\%) while remaining similar computational cost.
For a fair comparison with CoST~\cite{Cost}, we implement the architecture based on our baseline backbones by adding the CoST module into res4 and res5. Feeding 8-frame clips, as expected, our MVFNet-R50 is more lightweight than CoST-R50 (32.9G \emph{vs.} 46G). 
Table~\ref{tab:ablation:compare} shows the superiority of our MVFNet is quite impressive. 

\textbf{Deeper Backbone.}
In Table~\ref{tab:ablation:various} we compare various instantiations of MVFNet models on Kinetics-400. 
Thus far, all experiments used ResNet-50 as the backbone, we further study the performance of our MVFNet with a deeper backbone (\emph{i.e.}, ResNet-101 (R-101)). For models involving R-101, we use the same hyper-parameter settings as R-50 above.
As expected, using advanced backbones is complementary to our method. Comparing with the R-50 counterparts, our MVFNet gets additional improvement on R-101. We also investigate generalization of our models on longer input videos. Our models work well on longer sequences and all models have better results with longer inputs.

\textbf{Different Backbones.}
We further study the performance of MVFNet with MobileNet-V2 which is much smaller than the ResNet backbone. As shown in Table~\ref{tab:ablation:mbv2}, comparing with the 4frame-C2D, our 4frame-MVFNet achieves a better accuracy (67.5\% vs. 64.4\%) with the same MobileNet-V2 backbone on Kinetics-400.

\begin{table*}
	\centering
	\scalebox{0.875}{
      \begin{tabular}{ccccccc}
      \shline
      \multicolumn{1}{c}{\bfseries Method} & \multicolumn{1}{c}{\bfseries Backbone} & \multicolumn{1}{c}{\bfseries Frames$\times$Crops$\times$Clips} & \multicolumn{1}{c}{\bfseries FLOPs} & \multicolumn{1}{c}{\bfseries Pre-train} & \multicolumn{1}{c}{\begin{tabular}[c]{@{}c@{}}{\bfseries V1 Val} \\ {\bfseries Top-1 (\%)}\end{tabular}} &  \multicolumn{1}{c}{\begin{tabular}[c]{@{}c@{}}{\bfseries V2 Val} \\ {\bfseries Top-1 (\%)}\end{tabular}} 
      \\ \hline
      I3D (Wang et al. 2018) & 3D ResNet50 & \multirow{3}{*}{32\x3\x2} & \multicolumn{1}{c}{153G$\times$3$\times$2} & \multirow{3}{*}{\begin{tabular}[c]{@{}c@{}}ImageNet\\ +\\ K400\end{tabular}} & 41.6 & - \\
      NL I3D (Wang et al. 2018) & 3D ResNet50 &  & 168G\x3\x2 &  & 44.4 & -  \\
      NL I3D+GCN (Wang et al. 2018) & 3D ResNet50+GCN &  & 303G\x3\x2 &  & 46.1 & - \\ \hline

      ECO (Zolfaghari et al. 2018) & \multirow{2}{*}{BNIncep+3D Res18} & 8\x1\x1 & 32G\x1\x1 & \multirow{2}{*}{K400} & 39.6 & -  \\
      ECO$_{En}$ (Zolfaghari et al. 2018) &  & 92\x1\x1 & 267G\x1\x1 &  & 46.4 & - \\ \hline
      S3D-G~\cite{s3d} & Inception & 64\x1\x1  & 71G\x1\x1 & K400  & 48.2  & - \\   \hline
      \hline

      TSN \cite{tsn}  & ResNet50 & 8\x3\x2 & 33G\x3\x2 & ImageNet & 20.5 & 30.4 \\ \hline

      \multirow{2}{*}{TSM (Lin et al. 2019)} & \multirow{2}{*}{ResNet50} & 8\x3\x2 & 33G\x3\x2 & \multirow{2}{*}{\begin{tabular}[c]{@{}c@{}}ImageNet\end{tabular}} & 47.2 & 61.2 \\
       &  & 16\x3\x2 & 65G\x3\x2 &  & 48.4 & 63.1 \\
      \hline

     \multirow{2}{*}{STM \cite{stm}}  & \multirow{2}{*}{ResNet50} & 8\x3\x10 & 33G\x3\x10 & \multirow{2}{*}{ImageNet} & 49.2 & 62.3 \\
      &  & 16\x3\x10 & 67G\x3\x10 & & 50.7 & 64.3 \\   \hline

     \multirow{2}{*}{TEINet \cite{teinet}} & \multirow{2}{*}{ResNet50} & 8\x3\x10 & 33G\x3\x10 & \multirow{2}{*}{ImageNet} & 48.8 & 64.0 \\
      &  & 16\x3\x10 & 66G\x3\x10 & & 51.0 & 64.7 \\ \hline

     \multirow{2}{*}{TEA \cite{li2020tea}} & \multirow{2}{*}{ResNet50} & 8\x3\x10 & 35G\x3\x10 & \multirow{2}{*}{ImageNet} & 51.7 & - \\
      &  & 16\x3\x10 & 70G\x3\x10 & & 52.3 & - \\ \hline
      
       \multirow{5}{*}{MVFNet} & \multirow{5}{*}{ResNet50} & 8\x1\x1 & 33G\x1\x1 & \multirow{5}{*}{ImageNet} & 48.8 & 60.8  \\
       &  & 8\x3\x2 & 33G\x3\x2 & & 50.5 & 63.5 \\
      &  & 16\x1\x1 & 66G\x1\x1 & & 51.0 & 62.9  \\
       &  & 16\x3\x2 & 66G\x3\x2 & & \textbf{52.6} & \textbf{65.2} \\ 
       &  & (16+8)\x3\x2 & 99G\x3\x2 & & \textbf{54.0} & \textbf{66.3} \\ \hline      
      \shline
      \end{tabular}
	}
	\caption{Performance and FLOPs consumptions of our method on the Something-Something V1 and V2 datasets compared with the state-of-the-art methods. }
\label{some}
\label{tab:sota_sth}
\end{table*}

\subsection{Comparison with State-of-the-arts}
\textbf{Results on Kinetics-400.}
We make a comprehensive comparison in Table~\ref{tab:sota_kinetics}, where our MVFNet outperforms the recent SOTA approaches on Kinetics-400. Here we only list the models using RGB as inputs to perform comparisons. 
Compared with 2D CNN based models, when utilizing 8 frames as input, our MVFNet outperforms these current state-of-the-art efficient methods (\emph{e.g.}, TSM, TEINet and TEA) with a clear margin (\textbf{76.0}\% \emph{vs.} 74.1\%/74.9\%/75.0\%). 
Compared with computationally expensive models, MVFNet-R50 with 8-frame outperforms NL I3D-R50 with 32-frame (\textbf{76.0}\% \emph{vs.} 74.9\%) but only uses \textbf{2.1\x} less GFLOPs, and MVFNet-R50 with 16-frame uses \textbf{4.2\x} less GFLOPs than NL I3D-R50 with 128-frame but achieves a better accuracy (\textbf{77.0}\% \emph{vs.} 76.5\%). Moreover, 8-frame MVFNet-R101 achieves a competitive accuracy (\textbf{77.4}\% \emph{vs.} 77.2\%, 77.4\%, 77.7\%) when using \textbf{3\x}, \textbf{5.7\x}, \textbf{6.7\x} less GFLOPs than SlowOnly, NL I3D and SmallBig respectively. 
To mimics two-steam fusion with two temporal rates as done in SlowFast, we perform score fusion over 16-frame MVFNet-R101 and 8-frame MVFNet-R101. Our MVFNet$_{En}$ obtains better performance than SlowFast (\textbf{79.1\%} \emph{vs.} 78.9\%), using less GFLOPs (188G \emph{vs.} 213G).
TPN-I3D-101 achieves top-1 accuracy of 78.9\% with 32-frame clips as input, its complexity is as high as 374 GFLOPs. Our MVFNet$_{En}$ obtains better performance than TPN-R101, using 2\x less GFLOPs (188G \emph{vs.} 374G).

\textbf{Results on Something-Something V1 \& V2.}
The Something-Something datasets are more complicated than Kinetics. The comparison of our solution against existing state-of-the-arts are list in Table~\ref{tab:sota_sth}. These methods can be divided into two categories as shown in the two parts of Table~\ref{tab:sota_sth}. The upper part presents the 3D CNN based methods including S3D-G, ECO and I3D+GCN models. When compared with these methods, the obtained result shows substantial improvements and the \#FLOPs of our model is much smaller than these models. The lower part is 2D CNN based methods. Compared with these lightweight models which also target at improving efficiency, our solution yields a superior accuracy of 52.6\% on Something-Something V1 validation set and 65.2\% on V2 with 16 frames as input. 
For readers’ reference, here we also report the results of ensemble the models using 16 frames and 8 frames as inputs.

\subsection{Transfer Learning on UCF-101 \& HMDB-51}
We also evaluate the performance of our method on UCF-101 and HMDB-51 to show the generalization ability of MVFNet on smaller datasets. We finetune our MVFNet with 16 frames as inputs on these two datasets using model pre-trained on Kinetics-400 and report the mean class accuracy over three splits. From Table \ref{t:UCFHMDB}, we see that our model shows a pretty transfer capability and the mean class accuracy is 96.6\% and 75.7\% on UCF-101 and HMDB-51, respectively. 
As shown in the middle part of Table~\ref{t:UCFHMDB}, our MVFNet outperforms these 2D CNN based lightweight models. When comparing with the state-of-the-art models based on 3D convolutions such as I3D, R(2+1)D and S3D, our proposed MVFNet also obtains comparable or better performance.

\begin{table}
\centering
\scalebox{0.90}{
\begin{tabular}{cccc}
\shline
\textbf{Method} & \textbf{Backbone}  & \textbf{UCF-101}  & \textbf{HMDB-51} \\ \hline

ECO$_{En}$   & BNIncep+Res3D-18 & 94.8\%  & 72.4\% \\ 
ARTNet & ResNet-18 & 94.3\% & 70.9\% \\  
I3D   & Inception V1 & 95.6\%  & 74.8\% \\ 
R(2+1)D   & Inception V1 & 96.8\%  & 74.5\% \\
S3D-G & Inception V1 & 96.8\%  & 75.9\% \\ \hline

TSN & BNInception & 91.1\%  & - \\
StNet  &  ResNet-50 & 93.5\%  & - \\ 


TSM  & ResNet-50  & 95.9\%  & 73.5\%  \\
STM  & ResNet-50 & 96.2\%  & 72.2\%   \\ 
TEINet  & ResNet-50 & 96.7\%  & 72.1\%   \\ \hline

MVFNet  & ResNet-50 & \textbf{96.6}\%  & \textbf{75.7}\%   \\ 

\shline
\end{tabular}
}
\caption{\textbf{Mean class accuracy} on UCF-101 and HMDB-51 achieved by different methods which are transferred from their \textbf{Kinetics} models with RGB modality (over 3 splits).}
\label{t:UCFHMDB}
\end{table}

\section{Conclusion}
In this paper, we presented the Multi-View Fusion (MVF) module to better exploit spatiotemporal dynamics in videos. The MVF module works in a plug-and-play way and can be easily integrated into standard ResNet block to form a MVF block for constructing our MVFNet. We conducted a series of empirical studies to verify the effectiveness of MVFNet for video action recognition. Without any 3D convolution or pre-calculation of optical flow, the experimental results show that our method achieves the new state-of-the-art results on both temporal information dependent and spatial information dominated datasets with computational cost comparable to 2D CNN. In the future, we think the learnable $\beta$ to weight views will further boost the performance.

\bibstyle{aaai21}
\bibliography{references}
\end{document}